\documentclass[letterpaper, 10 pt, conference]{ieeeconf}  %
\usepackage{graphicx}
\usepackage{caption} %
\usepackage{etoolbox}
\usepackage{lipsum}
\usepackage{booktabs}
\usepackage{pifont}
\newcommand{\cmark}{\ding{51}}
\newcommand{\xmark}{\ding{55}}
\usepackage{diagbox}
\usepackage{multirow}
\usepackage{hyperref}
\usepackage[normalem]{ulem}
\usepackage{balance} %
\useunder{\uline}{\ul}{}

\usepackage{tabularx}
\newcolumntype{Y}{>{\centering\arraybackslash}X}
\newcolumntype{Z}{>{\raggedright\arraybackslash}X}

\newcommand{\greyrule}{\arrayrulecolor{black!30}\midrule\arrayrulecolor{black}}

\usepackage{nicematrix,tikz}
\usetikzlibrary{shadings}
\definecolor{synonymgreen}{rgb}{0.63, 0.87, 0.76}
\definecolor{depictionblue}{rgb}{0.60, 0.82, 0.91}
\definecolor{vissimyellow}{rgb}{0.94, 0.74, 0.42}
\definecolor{clutterpink}{rgb}{0.86, 0.64, 0.82}
\definecolor{missinggrey}{rgb}{0.82, 0.82, 0.82}
\definecolor{incorrectred}{rgb}{0.91, 0.51, 0.50}

\makeatletter
\apptocmd{\@maketitle}{\centering}{}{}%
\makeatother

\IEEEoverridecommandlockouts                              %

\title{\LARGE \bf
LEXI-SG: Monocular 3D Scene Graph Mapping with Room-Guided Feed-Forward Reconstruction
}

\author{Christina Kassab$^{*1}$, Hyeonjae Gil$^{*2}$, Mat{\'i}as Mattamala$^{3}$, Ayoung Kim$^{2}$, Maurice Fallon$^{1}$%
\thanks{$^{*}$ Indicates equal contribution}%
\thanks{$^{1}$ Christina Kassab and Maurice Fallon are with the Department of Engineering Science, University of Oxford, UK. Email: %
\texttt{\{christina, mfallon\}@robots.ox.ac.uk}}%
\thanks{$^{2}$ Hyeonjae Gil and Ayoung Kim are with the Department of Mechanical Engineering, Seoul National University, South Korea. %
Email: \texttt{\{h.gil, ayoungk\}@snu.ac.kr}} %
\thanks{$^{3}$ Mat{\'i}as Mattamala is with the School of Informatics, University of Edinburgh, UK. Email: \texttt{matias.mattamala@ed.ac.uk}} %
}

\begin{document}
\maketitle
\thispagestyle{empty}
\pagestyle{empty}

\begin{abstract}
   Scene graphs are becoming a standard representation for robot navigation, providing hierarchical geometric and semantic scene understanding. However, most scene graph mapping methods rely on depth cameras or LiDAR sensors. In this work, we present LEXI-SG, the first dense monocular visual mapping system for open-vocabulary 3D scene graphs using only RGB camera input. Our approach exploits the semantic priors of open-vocabulary foundation models to partition the scene into rooms, deferring feed-forward reconstruction to when each room is fully observed---enabling scalable dense mapping without sliding-window scale inconsistencies. We propose a room-based factor graph formulation to globally align room reconstructions while preserving local map consistency and naturally imposing the semantic scene graph hierarchy. Within each room, we further support open-vocabulary object segmentation and tracking. We validate LEXI-SG on indoor scenes from the Habitat-Matterport 3D and self-collected egocentric office sequences. We evaluate its performance against existing feed-forward SLAM methods, as well as established scene graphs baselines. We demonstrate improved trajectory estimation and dense reconstruction, as well as, competitive performance in open-vocabulary segmentation. LEXI-SG shows that accurate, scalable, open-vocabulary 3D scene graphs can be achieved from monocular RGB alone. Our project page and office sequences are available \href{https://ori-drs.github.io/lexisg-web/}{here}.
\end{abstract}

\section{INTRODUCTION}

3D scene graphs represent scenes sparsely and hierarchically---capturing both high-level semantic regions like rooms and individual objects. These representations offer several advantages for autonomous robotic systems: they support downstream tasks such as navigation \cite{Hovsg2024}, can operate in real-time \cite{ConceptGraphs2023}, and can scale to large environments \cite{Rosinol2021}. Recent work has further enhanced their generalizability by grounding open-vocabulary language models into scene graph representations \cite{ takmaz2023openmask3d, Maggio2024Clio}. 

However, current methods typically require high-quality depth and pose estimates~\cite{ConceptGraphs2023, Hovsg2024, takmaz2023openmask3d}, which presents a practical challenge in real-world deployments where such inputs may be noisy or unavailable. Constructing and maintaining accurate scene representations in real-world scenarios requires robust localization and mapping. Two decades of SLAM research has provided a foundation for this task \cite{ORBSLAM3_TRO, 8421746, DavisonRMS07}, with semantic SLAM systems extending this by leveraging scene semantics for pose estimation \cite{DBLP:journals/corr/abs-1808-08378, Salas-Moreno_2013_CVPR}. These methods require careful calibration procedures along with complex front-end engineering to achieve high performance and robustness in real settings.

\begin{figure}[t]
  \centering
  \includegraphics[width=\columnwidth]{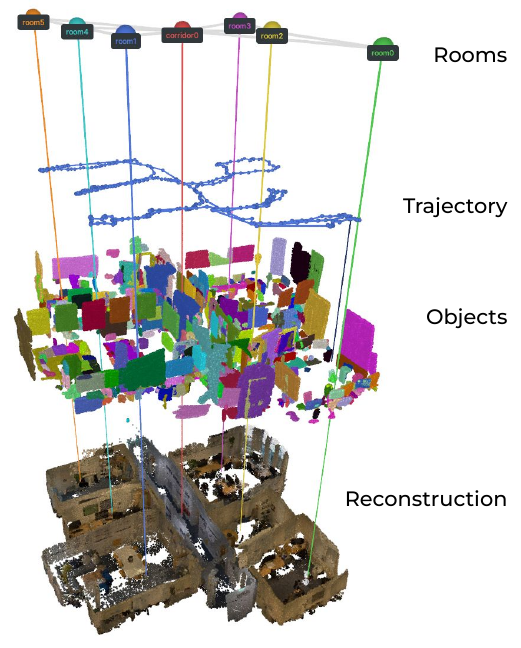} 
  \caption{\textbf{LEXI-SG is the first dense monocular mapping system to build open-vocabulary 3D scene graphs from RGB input alone.} We first partition the incoming image stream room by room. Within each room, we jointly estimate camera trajectories and dense geometry using feed-forward reconstruction models, amortizing expensive model queries while ensuring local scale consistency. The room graph is then expanded to a full 3D scene graph using open-vocabulary object segmentation.
\vspace{-1em}}
  \label{fig:scene-graph}
\end{figure}

\begin{figure*}[ht]
  \centering
  \includegraphics[width=\textwidth]{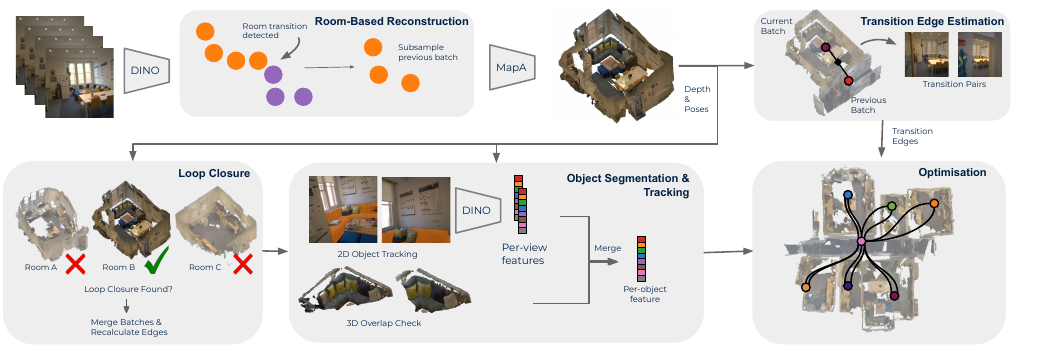} %
  \caption{\textbf{LEXI-SG System Overview.} RGB frames are segmented into rooms using DINO features. Upon detecting a room transition, the accumulated batch is passed through a feed-forward model (MapAnything---\textit{MapA} in the figure) to produce per-frame depths and poses in a local room frame. Transition edges are estimated by feeding transition frame pairs through the same model. New rooms are checked for loop closures, with any match triggering a merge. Once finalized, rooms are processed by the object segmentation module. The pose graph is then globally optimized over Sim(3).
  \vspace{-1em}
  }
  \label{fig:system-overview}
\end{figure*}

Recently, feed-forward reconstruction models \cite{wang2025vggt, mast3r_eccv24, keetha2026mapanything} have emerged as a new reconstruction  paradigm. By leveraging large sets of training samples, these models can infer dense reconstruction and pose estimates from uncalibrated camera setups without additional sensing. While their reconstruction capabilities in smaller scenes (e.g., rooms) is impressive and more flexible than engineered pipelines, scalability is limited. Recent work has sought to integrate these models into minimal ``plug-and-play'' SLAM systems to reconstruct larger scenes \cite{murai2024_mast3rslam, maggio2025vggt-slam, zhang2025vistaslam}. Integrating feed-forward SLAM methods with semantic scene representations remains largely unexplored, particularly at scale. 

In this work, we present LEXI-SG (Language EXtended Indoor Scene Graphs), a unified system for dense monocular mapping and semantic scene graph construction. Unlike current feed-forward SLAM systems, LEXI-SG defers reconstruction until each room is fully observed, amortizing model queries while improving both scale consistency and scene understanding. Our contributions are:

\begin{itemize}
    \item LEXI-SG, the first dense monocular SLAM system that builds an open-vocabulary 3D scene graph from RGB alone, without depth, or ground-truth pose input.
    \item A vision-only room identification method that detects transitions using DINO.
    \item A room-based reconstruction strategy that defers feed-forward inference until a room is fully observed, avoiding sliding-window scale inconsistencies.
    \item A Sim(3) room-level factor graph that globally aligns per-room reconstructions while preserving local consistency and correcting monocular scale ambiguity.
    \item An open-vocabulary segmentation module that lifts 2D mask tracklets into the scene graph as 3D object nodes.
    \item Evaluations on SLAM and scene graph tasks on standard datasets and self-collected office recordings.
\end{itemize}

\section{Related Work}

\subsection{3D Scene Understanding using Foundation Models}
Recent open-vocabulary 3D scene understanding methods have leveraged advances in visual foundation models such as CLIP~\cite{clip2021}, SAM~\cite{ravi2024sam} and DINO~\cite{simeoni2025dinov3}. These representations can be broadly categorized into two types: \textit{object-centric}~\cite{ConceptGraphs2023, Hovsg2024, lu2023ovir}, such as scene graphs, and \textit{dense}, which provide per-pixel or per-point semantic representations~\cite{Peng2023OpenScene, conceptfusion}.
    
Object-centric methods typically extract segments using instance segmentation techniques, either in 2D using models such as SAM~\cite{ConceptGraphs2023, Hovsg2024, lu2023ovir} or directly in 3D~\cite{takmaz2023openmask3d}. In the former case, segments are projected into 3D using depth images and temporally fused to create coherent scene representations. Spatial relationships between objects can be encoded as edges connecting object nodes, as demonstrated in ConceptGraphs~\cite{ConceptGraphs2023}. Other works extend this with higher-level layers such as rooms in HOV-SG~\cite{Hovsg2024}. Clio~\cite{Maggio2024Clio} optimizes efficiency by generating task-specific, compact scene graphs~\cite{Maggio2024Clio}.

In contrast dense methods assign semantic features to every 3D point, bypassing explicit object segmentation. OpenScene~\cite{Peng2023OpenScene} and ConceptFusion~\cite{conceptfusion} distill CLIP features into 3D point clouds, while LERF~\cite{lerf2023} embeds CLIP descriptors directly into a radiance field. Rayfronts~\cite{alama2025rayfronts} extends dense mapping beyond the depth range using semantic rays at map frontiers. While these methods offer fine-grained spatial resolution, they are often more computationally expensive compared to object-centric representations.

These methods typically rely on high-quality depth images which limits their use in real-world settings. LEXI-SG addresses this gap by constructing an open-vocabulary scene graph from monocular RGB input alone, leveraging feed-forward reconstruction in place of depth sensors and removing the dependency on ground truth poses. 

\subsection{Feed-Forward Monocular SLAM}
Visual SLAM has been studied for more than two decades with many generations of engineered methods proposed for camera motion estimation and 3D scene reconstruction. These approaches typically rely on feature extraction and tracking~\cite{ORBSLAM3_TRO} or the minimization of photometric error~\cite{engel14eccv} to estimate camera motion and reconstruct 3D scene structure. While effective in well-conditioned settings, they require careful calibration and engineering, and remain brittle under challenging illumination, motion blur, or low-texture scenes. 

Recent advances in feed-forward 3D reconstruction models, such as MASt3R~\cite{mast3r_eccv24}, VGGT~\cite{wang2025vggt} and MapAnything~\cite{keetha2026mapanything}, have demonstrated the ability to infer dense geometry and camera poses directly from multi-view inputs. This has motivated a growing body of research on SLAM systems that can operate without camera intrinsics yet can still generate dense scene representations. MASt3R-SLAM~\cite{murai2024_mast3rslam} couples MASt3R's two-view pointmap predictions with a tracking and global optimization back-end. Similarly, ViSTA-SLAM~\cite{zhang2025vistaslam} predicts per-pair dense pointmaps and relative poses using a symmetric two-view association network. In contrast, VGGT-based approaches~\cite{maggio2025vggt-slam, deng2025vggtlongchunkitloop} construct submaps in a rolling manner and use pose graph optimization to minimize scale and spatial drift between submaps.

While LEXI-SG also uses feed-forward reconstruction models, it does not use them on a sliding-window basis, but instead exploits semantic information across different levels to guide the reconstruction and pose estimation process. We demonstrate that using room-level semantics can improve the reconstruction results, refine the pose estimates, and enable scalable scene graph construction.

\subsection{Semantic SLAM Systems}
Semantic SLAM systems aim to tightly couple semantic and geometric information to jointly estimate camera poses and build scene representations. Early methods such as SLAM++~\cite{Salas-Moreno_2013_CVPR} and CubeSLAM~\cite{yang2019cubeslam} integrate object detections into the SLAM pipeline to jointly optimize camera poses and object landmarks. Inspired by these works, LEXI-SG extends the tight coupling of semantics and geometry to a monocular feed-forward setting. Room-level information is used to adaptively batch inputs for globally consistent reconstruction and to guide object segmentation and tracking, producing a full 3D scene graph without depth sensors or ground truth poses.

\section{Method}

\subsection{System Overview}
A system overview of LEXI-SG is presented in Fig~\ref{fig:system-overview}. The only input to the system is a stream of RGB images. The system maintains a room pose graph for optimization and incrementally populates a 3D scene graph with room and object nodes as output. The main modules of the system are: room-based reconstruction, object segmentation and tracking, loop closure, and global optimization. 

\subsection{Graph Structure}

\noindent\textbf{3D Scene Graph.}
The scene is represented as a hierarchical graph $\mathcal{G} = (\mathcal{V}, \mathcal{E})$ with two layers of nodes and
two types of edges. We denote nodes as $\mathcal{V} = \mathcal{V}_R \cup \mathcal{V}_O$. $\mathcal{V}_R$
is the set of room nodes and $\mathcal{V}_O$ is the set of object nodes. 

Each room node $v_{r_i} \in \mathcal{V}_R$ has an associated
local reference frame $\mathcal{F}_{r_i}$ with reference pose
$\mathbf{T}_{r_i} \in \mathrm{Sim}(3)$, and a
point cloud $\mathcal{P}_{r_i}$.  Each object node $v_{o_i} \in \mathcal{V}_O$ has an
associated frame $o_i$ with pose
$\mathbf{T}_{r_i o_i} \in \mathrm{Sim}(3)$ expressed \emph{relative to its parent room frame} $r_i$, a point cloud
$\mathcal{P}_{o_i}$, and a semantic feature vector $\mathbf{f}_{o_i}$. The edge set decomposes as
$\mathcal{E} = \mathcal{E}_{RR} \cup \mathcal{E}_{RO}$. Room-to-room
edges $e_{r_i r_j} \in \mathcal{E}_{RR}$ connect neighboring room nodes
and encode the relative transformation $\mathbf{T}_{r_i r_j}$ between
their local reference frames. Room-to-object edges
$e_{r_i o_i} \in \mathcal{E}_{RO}$ connect each object node to its
parent room node and encode the relative pose $\mathbf{T}_{r_i o_i}$
of the object frame in the room's local reference frame.

\noindent\textbf{Room Pose Graph.}
For global optimization we operate on a sub-graph of $\mathcal{G}$ obtained by restricting to the room layer,
\begin{equation}
  \mathcal{G}_P = (\mathcal{V}_R, \, \mathcal{E}_{RR}) \subseteq \mathcal{G},
\end{equation}
which we refer to as the \textit{room pose graph}. Its nodes are the
room reference poses $\mathbf{T}_{r_i}$ and its edges are the room-to-room relative transforms $\mathbf{T}_{r_i r_j}$, instantiated either as transition edges between temporally adjacent rooms
(Sec.~\ref{sec:transition_edge}) or as loop closure edges between revisited rooms (Sec.~\ref{sec:loop_closure}). The room pose graph is a Sim(3) pose graph and its optimization is described in Sec.~\ref{sec:global_optimization}.
Room-to-object edges $\mathcal{E}_{RO}$  encode parent--child containment in the hierarchy and are not optimized.

\begin{figure}[t]
  \centering
  \includegraphics[width=0.7\columnwidth]{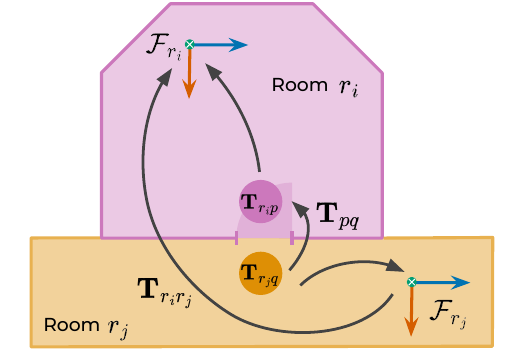} %
  \caption{\textbf{Transition edge estimation.} The relative transform $\mathbf{T}_{r_i r_j}$ between adjacent rooms is estimated by retrieving transition image pairs $(p, q)$ and computing $\mathbf{T}_{pq}$ via a feed-forward reconstruction model.
  \vspace{-1em}
  }
  \label{fig:transition-estimation}
\end{figure}

\begin{figure*}[ht]
  \centering
  \includegraphics[width=\textwidth]{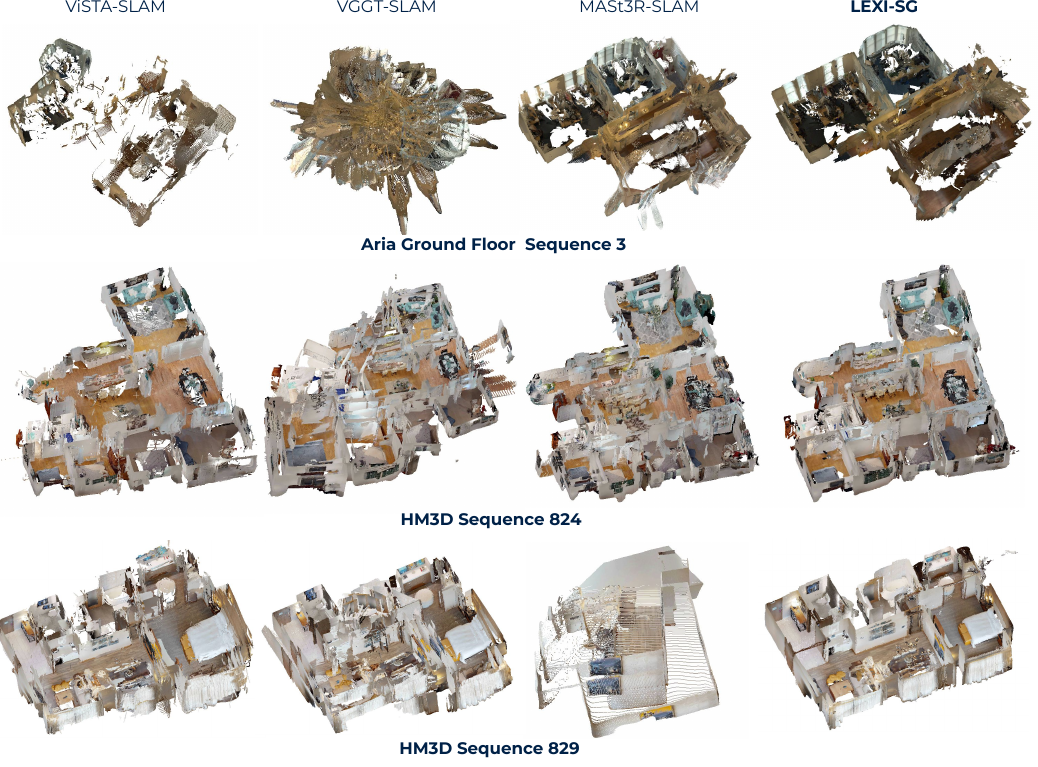} 
  \caption{\textbf{Qualitative reconstruction results on our office dataset AOD and HM3D.}
 LEXI-SG produces more accurate, globally consistent reconstructions with minimal double-walling and greater robustness across a range of indoor sequences.
 \vspace{-1em}
}
  \label{fig:reconstruction}
\end{figure*}

\subsection{Room-Based Reconstruction}
A common way to integrate feed-forward reconstruction models into a SLAM pipeline is to query them on a sliding window basis. However, this strategy produces locally inconsistent geometry across overlapping windows, leading to double-walled surfaces and scale drift---all of which degrades downstream object-level mapping. To avoid these limitations, we propose a room-aware strategy that defers reconstruction until a room has been fully observed, and then reconstructs it once from a single, curated batch of views. This results in a minimal but effective approach.

For each incoming frame, our system extracts DINO features $\mathbf{f}_t$, which are compared against text encodings of transition cues (e.g. `doorways' and `corridors') yielding a per-frame semantic label. Since these predictions can be noisy, room transitions are governed by a hysteresis mechanism: a running confidence score is accumulated across frames, and a room transition is only triggered when this score crosses a threshold. This requires there be a sustained signal to confirm a transition.

Upon detecting a room transition, the frames accumulated since the last transition are finalized as a batch. A small number of frames from the end of each batch are carried forward as an overlap into the following segment, ensuring continuity across rooms. The finalized batch is subsampled and passed through a feed-forward reconstruction model, which converts the images into a dense 3D point cloud together with per-frame depth estimates and poses defined relative to the local reference frame $\mathbf{T}_{r_i}$, anchored at the first image of the batch. This reference frame is subsequently incorporated as a node in the room pose graph. In this way, each room is reconstructed exactly once from its full set of observations, naturally reducing double-walling and scale inconsistencies inherent in sliding-window approaches. 

\subsection{Transition Edge Estimation}
\label{sec:transition_edge}
To construct an edge between two adjacent room nodes $v_i^r$ and $v_j^r$, we estimate the relative transformation $\mathbf{T}_{r_i r_j}$ between their respective local reference frames $\mathcal{F}_{r_i}$ and $\mathcal{F}_{r_j}$. As illustrated in Fig.~\ref{fig:transition-estimation}, this is achieved by retrieving a set of transition image pairs. We assume that the boundary between two rooms is observed in the final frames of the preceding batch and the initial frames of the following batch.

For each transition pair $(p, q)$, where $p$ and $q$ denote frame indices in $\mathcal{F}_{r_i}$ and $\mathcal{F}_{r_j}$ respectively, the two images are passed through the feed-forward reconstruction model to obtain their relative transform $\mathbf{T}_{pq}$. Here, $\mathbf{T}_{r_i p}$ denotes the pose of frame $p$ expressed in $\mathcal{F}_{r_i}$, and $\mathbf{T}_{r_j q}$ denotes the pose of frame $q$ expressed in $\mathcal{F}_{r_j}$. The transform between the two room reference frames is then recovered as:
\begin{equation}
    \mathbf{T}_{r_i r_j} = \mathbf{T}_{r_i p} \cdot \mathbf{T}_{pq} \cdot \mathbf{T}_{r_j q}^{-1}
\end{equation}
This process is repeated for all transition pairs, yielding a set of transform estimates $\{\mathbf{T}_{r_i r_j}\}$ which are stored as room-to-room edges $e_{r_i r_j} \in \mathcal{E}_{RR}$ in the room pose graph.

\begin{table*}[]
\centering
\caption{\textbf{Chamfer distance evaluation (m) on self-collected scenes from an office environment.} X denotes rooms for which a valid 
reconstruction could not be recovered. 
}
\resizebox{\textwidth}{!}{%
\begin{tabular}{@{}lccccccccccc@{}}
\toprule
           & \multicolumn{7}{c}{Floor 1}                                             & \multicolumn{4}{c}{Ground Floor}    \\
           & Room 0 & Room 1 & Room 2 & Room3 & Room 4      & Room 5      & Corridor & Room 0 & Room 1 & Room 2 & Corridor \\ \midrule
ViSTA-SLAM & 0.255  & X      & 0.365  & X     & {\ul 0.153} & {\ul 0.193} & 0.416    & X      & X      & X      & X        \\
VGGT-SLAM Sim(3)  & 0.215  & X      & X      & X     & X           & X           & X        & X      & X      & X      & X        \\
MASt3R-SLAM &
  {\ul 0.203} &
  {\ul 0.225} &
  {\ul 0.201} &
  {\ul 0.210} &
  X &
  X &
  {\ul 0.163} &
  {\ul 0.213} &
  {\ul 0.152} &
  \textbf{0.130} &
  {\ul 0.147} \\
LEXI-SG &
  \textbf{0.114} &
  \textbf{0.147} &
  \textbf{0.136} &
  \textbf{0.170} &
  \textbf{0.128} &
  \textbf{0.148} &
  \textbf{0.085} &
  \textbf{0.126} &
  \textbf{0.095} &
  {\ul 0.235} &
  \textbf{0.093} \\ \bottomrule
\end{tabular}%
}
\label{tab:chamf-dist}
\end{table*}

\subsection{Loop Closure}
\label{sec:loop_closure}
As the agent traverses the environment, it may revisit previously
observed rooms. To detect this, the loop closure module maintains a
database of room nodes encountered so far, and checks each newly
finalized room node against this database for potential matches.

For each new room node $v_{r_i}$, its image features are compared
against those of every room in the database using cosine similarity.
A candidate match $v_{r_j}$ is confirmed if the number of image
feature pairs with a similarity score above $\tau_s$ exceeds a
threshold $\tau_r$. Upon finding a candidate match, the image sets of the two nodes are
merged to form a unified room batch. This merged batch is passed
through the feed-forward reconstruction model to produce a new set of
local poses and a unified room reconstruction, yielding a candidate
node $\tilde{v}_r$ with local reference frame $\mathcal{F}_{\tilde{r}}$
that replaces both $v_{r_i}$ and $v_{r_j}$ in the room pose graph.

Before finalizing the merge, the geometric consistency of the
candidate node with its neighbors must be verified. For each room node
$v_{r_k}$ that shares a room-to-room edge with either $v_{r_i}$ or
$v_{r_j}$, a new relative transformation $\mathbf{T}_{\tilde{r} r_k}$
is estimated via pairwise boundary frame matching, as described in
Sec.~\ref{sec:transition_edge}. The merge is only accepted if valid
transformations can be estimated for \emph{all} affected edges.

If the merge is accepted, the two original nodes are removed from the
room pose graph and replaced by the candidate node $\tilde{v}_r$. All
room-to-room edges $\mathcal{E}_{RR}$ previously incident to $v_{r_i}$
or $v_{r_j}$ are deleted and replaced by the newly verified loop
closure edges connecting $\tilde{v}_r$ to its neighbors. The candidate
node is then added to the database, and the original entries are
removed. If no match is found, or if edge verification fails, the new node is
added to the database unchanged and the graph remains unmodified.

\subsection{Open-Vocabulary Object Segmentation and Tracking}

Once a room has been finalized and any loop closures resolved, its image frames (drawn from the room's existing batch) are passed to the object segmentation and tracking module, which populates the room with child object nodes to complete the scene graph hierarchy.

Given RGB-D sequences for each room processed with a feed-forward reconstruction model, per-frame object masks are generated using Recognize Anything~\cite{zhang2024recognize} and Grounding Dino~\cite{liu2024grounding}. We use a SAM2-based \cite{ravi2024sam} tracking module to propagate each seed mask to adjacent frames and form object-centric tracklets across views. This propagation step turns an originally single-view signal into a multi-view signal, which provides a stronger support for subsequent 3D object reconstruction. It also makes cross-view mask association possible directly in image space through mask-level overlap, not solely relying on the 3D point cloud overlap. The propagated and merged mask tracklets are lifted into object-level point clouds. Each object point cloud becomes an object node $v_{o_i} \in \mathcal{V}_{O}$ with pose $\mathbf{T}_{r_i o_i}$ expressed in the parent room frame and attached via a room-to-object edge $e_{r_i o_i} \in \mathcal{E}_{RO}$ completing the scene graph hierarchy.

\subsection{Global Optimization}
\label{sec:global_optimization}
We globally optimize the room pose graph $\mathcal{G}_P = (\mathcal{V}_R, \mathcal{E}_{RR})$ over the $\mathrm{Sim}(3)$ Lie group to minimize inconsistencies introduced by accumulated drift and noisy pairwise estimates, allowing the optimizer to correct for the scale ambiguity inherent in monocular
reconstruction. Only the room reference poses $\mathbf{T}_{r_i}$ are optimized; object poses $\mathbf{T}_{r_i o_i}$, stored in room-local frames, update implicitly, so the full 3D scene graph remains
globally consistent. The
resulting factor graph is solved using the Levenberg--Marquardt optimizer.

\section{Experiments}

\begin{table}[]
\centering
\caption{\textbf{Absolute trajectory error (ATE (m)) on selected scenes from Habitat-Matterport 3D.} X denotes failed sequences or sequences with more than 2m of error.}
\footnotesize
\resizebox{\columnwidth}{!}{%
\begin{tabular}{@{}lccccccc@{}}
\toprule
 & 824 & 829 & 843 & 847 & 873 & 877 & 890 \\ \midrule
ViSTA-SLAM       & {\ul 0.351}    & {\ul 0.309}    & {\ul 1.154}    & 1.884          & X              & {\ul 0.576}    & X              \\
VGGT-SLAM SL(4)  & X              & X              & X              & X              & X              & X              & X              \\
VGGT-SLAM Sim(3) & 1.153          & 0.849          & 1.777          & 1.431          & X              & 0.872          & {\ul 0.613}    \\
VGGT-SLAM 2      & 0.703          & 0.611          & \textbf{0.620} & X              & \textbf{0.477} & 0.955          & \textbf{0.553} \\
MASt3R-SLAM      & 0.693          & X              & 1.521          & {\ul 0.717}    & X              & X              & 1.695          \\
LEXI-SG          & \textbf{0.343} & \textbf{0.143} & {\ul 0.628}    & \textbf{0.461} & X              & \textbf{0.554} & 0.712          \\ \bottomrule
\end{tabular}%
}
\label{tab:hm3d-ate}
\end{table}

\begin{table}[]
\centering
\caption{\textbf{Absolute trajectory error (ATE (m)) from the Aria office environment.} X denotes failed sequences or sequences with more than 2m of error.
}
\resizebox{\columnwidth}{!}{%
\begin{tabular}{@{}lccccccc@{}}
\toprule
                 & \multicolumn{3}{c}{Floor 1}                                 & \multicolumn{3}{c}{Ground Floor}      & \multicolumn{1}{c}{} \\
                 & Seq 1                     & Seq 2          & Seq 3          & Seq 1          & Seq 2          & Seq 3                 & avg                  \\ \midrule
ViSTA-SLAM       & 0.840                     & 1.038          & 1.405          & 0.668          & 0.291          & 1.651                 & {\ul 0.982}          \\
VGGT-SLAM SL(4)  & \multicolumn{1}{c}{1.194} & X              & X              & 0.745          & 0.804          & \multicolumn{1}{c}{X} & -                    \\
VGGT-SLAM Sim(3) & 1.320                     & 0.751          & 0.469          & 0.787          & 0.503          & \multicolumn{1}{c}{X} & -                    \\
VGGT-SLAM 2      & 1.037                     & {\ul 0.586}    & X              & {\ul 0.548}    & 0.485          & \multicolumn{1}{c}{X} & -                    \\
MASt3R-SLAM      & {\ul 0.302}               & X              & {\ul 0.294}    & \textbf{0.241} & \textbf{0.139} & {\ul 0.296}           & -                    \\
LEXI-SG          & \textbf{0.166}            & \textbf{0.277} & \textbf{0.277} & 0.647          & {\ul 0.201}    & \textbf{0.262}        & \textbf{0.305}       \\ \bottomrule
\end{tabular}%
}
\label{tab:ori-ate}
\end{table}

\subsection{Experimental Setup}
We validated LEXI-SG across four standard SLAM and scene graph tasks: Pose Estimation (Task 1), Dense Reconstruction (Task 2), Room Segmentation (Task 3), and Open-Vocabulary Object Segmentation (Task 4). 

All experiments are run on a workstation with an NVIDIA RTX 4090 GPU.  We provide the technical details of the LEXI-SG configuration, baselines, and datasets as follows. 

\noindent\textbf{LEXI-SG.} We use MapAnything~\cite{keetha2026mapanything} as the feed-forward reconstruction model throughout, with a batch size of 60 to ensure sufficient frame coverage across larger rooms and corridors. MapAnything was selected over models such as VGGT and DepthAnything3 despite its marginally lower pose estimation accuracy because our experiments show that it is more robust in long corridors. However, LEXI-SG could be easily adapted to other similar models. 

\noindent\textbf{Baselines.} For Camera Pose Estimation (Task 1) and Dense Reconstruction (Task 2) we compared against MASt3R-SLAM~\cite{murai2024_mast3rslam}, the Sim(3) 
and SL(4) variants of VGGT-SLAM~\cite{maggio2025vggt-slam}, VGGT-SLAM2~\cite{maggio2025vggt-slam2} and 
ViSTA-SLAM~\cite{zhang2025vistaslam}.

For Room Segmentation (Task 3) we compared against HOV-SG~\cite{Hovsg2024} and Hydra~\cite{Hughes2022}. For Open-Vocabulary Object Segmentation (Task 4) we compared against comparable object-centric methods such as ConceptGraphs~\cite{ConceptGraphs2023}.

\noindent\textbf{Datasets.} For pose estimation and reconstruction (Tasks 1 and 2), we evaluate on multi-room datasets rather than single-room benchmarks such as TUM RGB-D~\cite{sturm12iros}, since a single room constitutes one batch and would evaluate the feed-forward model alone rather than our full system.

Therefore, we evaluate on sequences from Habitat-Matterport 3D (HM3D)~\cite{ramakrishnan2021hm3d}, a large-scale dataset of multi-room, multi-floor indoor home environments. Furthermore, we used self-collected sequences from a two-floor office environment recorded 
using Meta's Project Aria glasses (Gen 1) which we call the \textit{Aria Office Dataset} (AOD).  AOD contains six sequences in total, three per floor, each spanning between 2 and 6 rooms and including loop closures. We treat the output poses from the Meta's  multi-camera visual-inertial SLAM system as ground-truth. Although atypical, this system has been shown to achieve sub-centimetre accuracy in indoor settings~\cite{Krishnan_2025_ICCV}, providing reliable ground truth. We use only the rectified forward-facing RGB camera with an FoV of $110^{\circ}$ from the glasses in our experiments.

The semantic evaluations (Tasks 3 and 4) were performed using the protocol defined in OpenLex3D~\cite{kassab2025openlex3d}.

\begin{table}[]
\centering
\caption{\textbf{Room segmentation evaluation.} Precision and recall are calculated based on the metric
provided by Hydra~\cite{Hughes2022}. Hydra, HOV-SG and LEXI-SG GT use ground truth poses and RGB-D images in this experiment.
}
\resizebox{\columnwidth}{!}{%
\begin{tabular}{@{}ccccccccc@{}}
\toprule
\multicolumn{1}{l}{} &
  \multicolumn{1}{l}{} &
  \multicolumn{1}{r}{824} &
  \multicolumn{1}{r}{829} &
  \multicolumn{1}{r}{843} &
  \multicolumn{1}{r}{873} &
  \multicolumn{1}{r}{877} &
  \multicolumn{1}{r}{890} &
  \multicolumn{1}{l}{Avg} \\ \midrule
\multirow{4}{*}{Precision} & LEXI-SG                        & 0.52 & 0.58 & 0.77 & X    & 0.54 & 0.62 & \multicolumn{1}{c}{-} \\
                           & \multicolumn{1}{l}{LEXI-SG GT} & 0.56 & 0.75 & 0.77 & 0.78 & 0.59 & 0.81 & 0.71                  \\
                           & HOV-SG                         & 0.81 & 0.86 & 0.88 & 0.95 & 0.74 & 0.94 & {\ul 0.86}            \\
                           & Hydra                          & 0.79 & 0.88 & 0.87 & 0.96 & 0.81 & 0.95 & \textbf{0.88}         \\ \midrule
\multirow{4}{*}{Recall}    & LEXI-SG                        & 0.85 & 0.98 & 0.69 & X    & 0.86 & 0.74 & \multicolumn{1}{c}{-} \\
                           & \multicolumn{1}{l}{LEXI-SG GT} & 0.96 & 0.91 & 0.74 & 0.85 & 0.78 & 0.93 & \textbf{0.86}         \\
                           & HOV-SG                         & 0.80 & 0.88 & 0.87 & 0.67 & 0.92 & 0.87 & {\ul 0.84}            \\
                           & Hydra                          & 0.78 & 0.85 & 0.78 & 0.80 & 0.88 & 0.62 & 0.79                  \\ \bottomrule
\end{tabular}%
}
\label{tab:room-seg}
\end{table}

\subsection{Task 1 -- Camera Pose Estimation}

Results for HM3D and AOD sequences are presented in 
Tabs.~\ref{tab:hm3d-ate} and~\ref{tab:ori-ate} respectively. LEXI-SG 
achieves the lowest average trajectory error across the datasets. On HM3D, 
VGGT-SLAM2 achieves the next best performance, though all monocular methods 
struggle on the larger-scale sequences. The SL(4) variant of VGGT-SLAM regularly diverges on HM3D, with Sim(3) proving more stable. Our room-based image batching approach yields improvements on these sequences, demonstrating how monocular feed-forward SLAM can be scaled to large environments.

On the AOD sequences, MASt3R-SLAM achieves the second best performance. The performance
of LEXI-SG on 
ground floor sequences 1 and 2 is reduced by some room segmentation 
errors, where portions of a large meeting room are misclassified as corridor. 
As shown in Fig.~\ref{fig:reconstruction}, improved pose estimation translates 
directly to more globally consistent reconstructions, with minimal 
double-walling and reduced overlap between adjacent rooms. 

We attribute these gains to our room-based reconstruction: deferring feed-forward inference until a room is fully traversed gives each batch maximal co-visibility, improving per-room reconstruction quality and reducing the drift that accumulates when chaining many sliding-window batches.

\begin{figure}[t]
  \centering
  \includegraphics[width=\columnwidth]{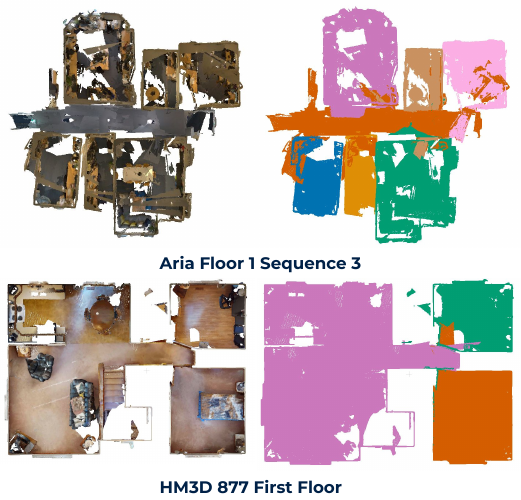} 
  \caption{\textbf{Qualitative room segmentation results on the AOD and HM3D sequences.} Our approach reliably delineates room boundaries by detecting transitional structures (such as doorways and corridors) from RGB input alone---without relying on a depth sensor or geometric priors.
  }
  \label{fig:room-segmentation}
  \vspace{-2em}
\end{figure}

\subsection{Task 2 -- Dense Reconstruction}
We segment the reconstructed scenes from each method into individual rooms and evaluate the resulting geometry against ground-truth terrestrial laser scans using the AOD sequences. We use the same set of baselines listed above.

To evaluate reconstruction accuracy we present chamfer distance results in Tab.~\ref{tab:chamf-dist}. We calculate the chamfer distances in the same manner as MASt3R-SLAM~\cite{murai2024_mast3rslam}. We denote with an X any rooms for which a valid reconstruction could not be recovered. On average LEXI-SG achieves the lowest reconstruction error, with MASt3R-SLAM achieving the next best results. ViSTA-SLAM and VGGT-SLAM produced reconstructions with overlapping rooms, though a small number of rooms could still be extracted and evaluated. VGGT-SLAM2 was excluded, as it either produced overlapping rooms or failed to converge. 

Fig~\ref{fig:reconstruction} shows qualitative comparisons with baseline methods. Our room-based batching approach achieves greater local as well as global geometric consistency. These improvements stem from LEXI-SG reconstructing each room in a single pass. In contrast, sliding-window approaches must fuse many partial reconstructions from overlap windows, blurring fine geometric detail.

\begin{figure}[]
  \centering
  \includegraphics[width=\columnwidth]{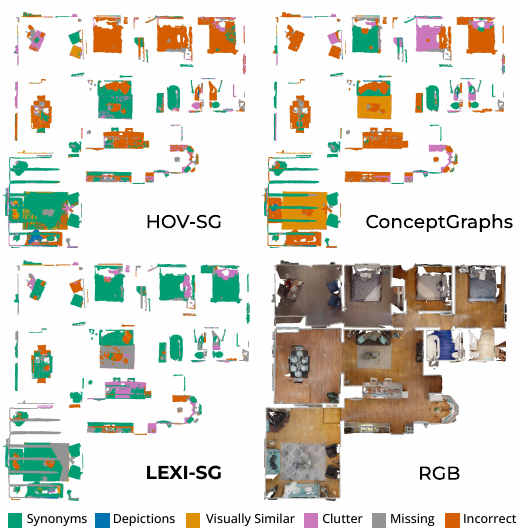}
  \caption{\textbf{Visualization of OpenLex3D Benchmark on 00824 sequence of HM3D dataset using ground-truth poses and depth.} Our object segmentation module shows better performance in the synonyms category (green) and fewer incorrect labels compared to other baselines.
  }
  \label{fig:object-segmentation}
  \vspace{-2em}
\end{figure}

\subsection{Task 3 -- Room Segmentation} 

We evaluate on HM3D sequences following the evaluation protocol of 
Hydra~\cite{Hughes2022}, and compare against HOV-SG and 
Hydra. These methods use ground truth poses and depth images for their evaluations.

We present quantitative and qualitative room segmentation results in Tab.~\ref{tab:room-seg} and Fig.~\ref{fig:room-segmentation}, respectively. 
While relying only on RGB image, LEXI-SG achieves comparable recall, albeit with lower precision. This precision gap can be attributed to our room transition-based approach: by detecting room boundaries through doorways and corridors, our method does not fully segment open-plan layouts present in the HM3D sequences, as illustrated in the bottom example of Fig.~\ref{fig:room-segmentation}. 

We also evaluate LEXI-SG using the ground truth poses and the depth images provided with HM3D. In this setting, both precision and recall improve. LEXI-SG GT surpasses HOV-SG and Hydra in recall, while precision increases but remains comparatively lower, again owing to the limitations in our method for room segmentation in open-plan spaces. Nevertheless, the results demonstrate that reasonably accurate room segmentation is achievable from monocular RGB alone, without access to ground-truth poses or depth. 

\subsection{Task 4 -- Open-Vocabulary Semantic Segmentation}

We evaluate our object segmentation module on the OpenLex3D benchmark~\cite{kassab2025openlex3d}, using its semantic segmentation metric. Results are reported in Tab.~\ref{tab:iou-top5}. Following the same protocol as used by ConceptGraphs, HOV-SG, and OpenMask3D, we use the provided ground truth poses and depth images for each segment. This allows us to isolate the semantic segmentation performance, ensuring that the evaluation is not affected by potential misalignments between predicted and ground-truth geometry. A perfect score corresponds to a similarity (S) of 1.0, and 0.0 across all remaining categories. 

On the Replica and ScanNet++ datasets, LEXI-SG achieves performance comparable to HOV-SG, and achieves the best synonym score on HM3D. These results suggest that tracking segments in 2D is more favourable than 3D merging strategies based on IoU and cosine similarity thresholds, as employed by ConceptGraphs and OpenMask3D. The strong performance of HOV-SG suggests that DBSCAN feature clustering yields more discriminative embeddings than averaging; however, in our approach we opt for averaging as it avoids additional computational overhead. 

\begin{table}[]
\centering
\caption{\textbf{Semantic Segmentation Results on the OpenLex3D Benchmark (using ground-truth poses and depth).} \textit{S} is the FREQ at synonyms, \textit{D} is depictions, \textit{VS} is visually similar, \textit{C} is clutter, \textit{M} is missing and \textit{I} is incorrect. 
}
\resizebox{\columnwidth}{!}{%
\begin{NiceTabular}{clcccccc}
\CodeBefore
    \columncolor{synonymgreen!50}{3}
    \columncolor{incorrectred!50}{8}
\Body
\toprule
\textbf{Data} &
  \textbf{Method} &
\multicolumn{1}{l}{\Block[tikz={left color=synonymgreen, right color=synonymgreen}]{1-1}{} $S$ ↑} &
\multicolumn{1}{l}{\cellcolor{depictionblue}$D$ ↓} &
\multicolumn{1}{l}{\cellcolor{vissimyellow}$VS$ ↓} &
\multicolumn{1}{l}{\cellcolor{clutterpink}$C$ ↓} &
\multicolumn{1}{l}{\cellcolor{missinggrey}$M$ ↓} &
\multicolumn{1}{l}{\Block[tikz={left color=incorrectred, right color=incorrectred}]{1-1}{} $I$ ↓} \\ \midrule
                            & ConceptGraphs~\cite{ConceptGraphs2023}          & 0.41          & {\ul 0.01}    & 0.11          & {\ul 0.24}    & \textbf{0.02} & 0.22          \\
                            & HOV-SG~\cite{Hovsg2024}                         & \textbf{0.45} & \textbf{0.00} & \textbf{0.05} & 0.27          & {\ul 0.07}    & {\ul 0.16}    \\
                            & OpenMask3D~\cite{takmaz2023openmask3d}          & {\ul 0.43}          & 0.01 & {\ul 0.07}    & 0.29 & 0.10          & \textbf{0.10}    \\
\multirow{-4}{*}{\rotatebox{90}{Replica}}   & LEXI-SG                        & 0.42    & \textbf{0.00} & {\ul 0.07}          & \textbf{0.19}    & 0.16          & {\ul 0.16} \\ \greyrule
                            & ConceptGraphs~\cite{ConceptGraphs2023}          & 0.26          & {\ul 0.02}    & 0.05          & \textbf{0.10} & 0.13          & 0.44          \\
                            & HOV-SG~\cite{Hovsg2024}                         & \textbf{0.40} & {\ul 0.02}    & {\ul 0.04}    & {\ul 0.16}    & \textbf{0.08} & 0.30    \\
                            & OpenMask3D~\cite{takmaz2023openmask3d}          & 0.27          & \textbf{0.01} & \textbf{0.03} & 0.29          & 0.13          & {\ul 0.27} \\
\multirow{-4}{*}{\rotatebox{90}{ScanNet++}}  & LEXI-SG                       & \textbf{0.40}    & \textbf{0.01} & \textbf{0.04} & 0.19          & {\ul 0.12}    & \textbf{0.24} \\ \greyrule
                            & ConceptGraphs~\cite{ConceptGraphs2023}          & 0.27          & {\ul 0.02}    & \textbf{0.03} & \textbf{0.12} & \textbf{0.08} & 0.47          \\
                            & HOV-SG~\cite{Hovsg2024}                         & {\ul 0.33}    & {\ul 0.02}    & {\ul 0.04}    & 0.18          & \textbf{0.08} & {\ul 0.36}    \\
                            & OpenMask3D~\cite{takmaz2023openmask3d}          & 0.31          & \textbf{0.01} & \textbf{0.03} & {\ul 0.13}    & 0.26          & \textbf{0.26} \\
\multirow{-4}{*}{\rotatebox{90}{HM3D}}      & LEXI-SG                        & \textbf{0.41} & \textbf{0.01} & \textbf{0.03} & 0.14          & {\ul 0.16}    & \textbf{0.26} \\ \bottomrule
\end{NiceTabular}%
}
\label{tab:iou-top5}
\end{table}

\begin{table}[b]
\centering
\caption{\textbf{Ablation results reporting average ATE (m) on AOD and HM3D (excluding seq 877).} Room-based reconstruction improves performance on both datasets, with the loop closure module yielding further improvements. }
\begin{tabular}{cc|cc}
\toprule
\shortstack{\textbf{Room-Based} \\ \textbf{Reconstruction}} & \shortstack{\textbf{Loop} \\ \textbf{Closure}} & \textbf{AOD} & \textbf{HM3D} \\
\midrule
\xmark & \xmark & 0.812 & 0.626 \\
\cmark & \xmark & \underline{0.427} & \underline{0.518} \\
\cmark & \cmark & \textbf{0.305} & \textbf{0.474} \\
\bottomrule
\end{tabular}

\label{tab:ablation}
\end{table}

\begin{table}[]
\centering
\caption{\textbf{Ablation on batch size showing average ATE (m) on AOD and HM3D.} A batch size of 60 is best on both datasets; AOD is relatively insensitive to batch size while smaller batches degrade HM3D performance.
}
\begin{tabular}{c|cc}
\toprule
\textbf{Batch Size} & \textbf{AOD} & \textbf{HM3D} \\
\midrule
30 & 0.467 & 1.074 \\
60 & \textbf{0.305} & \textbf{0.474} \\
90 & \underline{0.401} & \underline{0.594} \\
\bottomrule
\end{tabular}
\label{tab:batch-size}
\end{table}

\begin{table}[h]
\centering
\caption{\textbf{Runtime Summary for AOD seq 3 (4248 frames).} The system runs at \textbf{12.92 FPS} without object segmentation and \textbf{1.56 FPS} with.}
\begin{tabular}{lrr}
\hline
\textbf{Stage} & \textbf{Time (s)} & \textbf{Share (\%)} \\
\hline
Feature extraction      & 35.63   & 1.31  \\
Model inference         & 126.69  & 4.65  \\
PCD building            & 3.87    & 0.14 \\
Transition pairs        & 34.83   & 1.28  \\
Loop closure (semantic) & 31.27   & 1.15  \\
Loop closure (IoU)      & 96.54   & 3.55  \\
Object segmentation     & 2394.12 & 87.9  \\
Optimisation            & 0.01    & 0.00 \\
\hline
Total w/o objects       & 328.84  &       \\
Total w/ objects        & 2722.96 & 100.0 \\
\hline
\end{tabular}
\label{tab:runtime}
\end{table}

\subsection{Ablations}

Tables~\ref{tab:ablation} and~\ref{tab:batch-size} report ATE ablations for the main components of LEXI-SG. Room-based reconstruction nearly halves ATE on AOD over a sliding-window baseline, with a smaller but consistent improvement on HM3D, confirming that deferring feed-forward inference until a room is fully observed reduces drift from chaining overlapping windows. Loop closures further reduce error, with a smaller effect on HM3D due to fewer revisited rooms

Tab.~\ref{tab:batch-size} reports ATE across varying batch sizes. A batch size of 60 performs best: AOD is relatively insensitive to this choice, while smaller batches degrade HM3D performance as they fail to capture sufficient coverage of larger rooms within a single pass. A batch of 60 also allows the retention of enough frames to support reliable object tracking.

Table~\ref{tab:runtime} breaks down the per-stage runtime of the pipeline for an AOD sequence on a NVIDIA RTX PRO 6000, showing that object segmentation dominates the total cost at 87.9\%, followed by MapAnything inference at 4.65\%. The full pipeline runs at 12.92 FPS without object segmentation and at 1.56 FPS with object segmentation.

\section{conclusion}
We presented LEXI-SG, a monocular SLAM system that tightly couples feed-forward reconstruction with semantic scene graph mapping, achieving globally consistent reconstructions and competitive semantic understanding from RGB input alone. Evaluations across pose estimation, dense reconstruction, room segmentation, and open-vocabulary object segmentation demonstrate state-of-the-art performance among monocular feed-forward SLAM methods. The two primary limitations of the system are the reduced room segmentation precision in open-plan spaces, where the absence of doorways or corridors prevents transition detection, and a pose accuracy ceiling imposed by the feed-forward model, since keyframes are not individually optimized in order to preserve local batch consistency. Future work could address the former by incorporating geometric or appearance-based cues to handle open-plan layouts.

\section*{ACKNOWLEDGMENT}
The work at the University of Oxford was supported by a Royal Society University Research Fellowship (Fallon, Kassab), and the work at Seoul National University (Kim, Gil) is supported by the National Research Foundation of Korea (NRF) grant funded by the Korea government (MSIT)(No. RS-2024-00461409).

\balance
\bibliographystyle{IEEEtran}
\bibliography{references}

\end{document}